%% file: main.tex
\documentclass[letterpaper]{article} 
\usepackage{aaai24}  
\usepackage{times}  
\usepackage{helvet}  
\usepackage{courier}  
\usepackage[hyphens]{url}  
\usepackage{graphicx} 
\usepackage{natbib}  
\usepackage{caption} 
\usepackage{algorithm}
\usepackage{algpseudocode} 
\usepackage{graphicx}
\usepackage{mathtools}  
\usepackage{amsmath}
\usepackage{amssymb}
\usepackage{tabulary}
\usepackage{booktabs}

\usepackage{newfloat}
\usepackage{listings}
\DeclareCaptionStyle{ruled}{labelfont=normalfont,labelsep=colon,strut=off} 
\floatstyle{ruled}
\newfloat{listing}{tb}{lst}{}
\floatname{listing}{Listing}
\usepackage{color}

\usepackage{subcaption}

\title{Promoting Research Collaboration with Open Data Driven Team Recommendation in Response to Call for Proposals}

\author {
    Siva Likitha Valluru\textsuperscript{\rm 1},
    Biplav Srivastava\textsuperscript{\rm 1},
    Sai Teja Paladi\textsuperscript{\rm 1},
    Siwen Yan\textsuperscript{\rm 2},
    Sriraam Natarajan\textsuperscript{\rm 2}
}
\affiliations {
    \textsuperscript{\rm 1}Artificial Intelligence Institute, University of South Carolina\\
    \textsuperscript{\rm 2} Statistical Artificial Intelligence and Relational Learning Group (StARLinG Lab), University of Texas\\
    \{svalluru@email., 
    biplav.s@, spaladi@email.\}sc.edu, \{siwen.yan, sriraam.natarajan\}@utdallas.edu
}

\begin{document}

\maketitle

\begin{abstract}
Building teams and promoting collaboration 
are two very common business activities. An example of these are seen in the {\em TeamingForFunding} problem, where research institutions and researchers are interested to identify collaborative opportunities when applying to funding agencies in response to latter's calls for proposals. We describe a novel {\em deployed} system to recommend teams using a variety of AI methods, such that (1) each team achieves the highest possible  skill coverage that is demanded by the opportunity, and (2) the workload of distributing the opportunities is balanced amongst the candidate members. We address these questions by extracting skills latent in open data of proposal calls (demand) and researcher profiles (supply), normalizing them using taxonomies, and creating efficient algorithms that match demand to supply. We create teams to maximize goodness along a novel metric balancing short- and long-term objectives. We validate the success of our algorithms (1)~quantitatively, by evaluating the recommended teams using a goodness score and find that more informed methods lead to recommendations of smaller number of teams but higher goodness, and (2)~qualitatively, by conducting a large-scale user study at a college-wide level, and demonstrate that users overall found the tool very useful and relevant. Lastly, we evaluate our system in two diverse settings in US and India (of researchers and proposal calls) to establish generality of our approach, and deploy it at a major US university for routine use.

\end{abstract}


\input{sections/introduction}
\input{sections/relatedwork}
\input{sections/problem}
\input{sections/systemdesign}
\input{sections/evaluation}

\input{sections/discussion}

\input{sections/conclusion}
\input{sections/acknowledgement}

\bigskip

\bibliography{references}

\end{document}

%% file: sections/introduction.tex
\section{Introduction}

In the recent decade, there has been an increased interest in studying teamwork skills and their impacts in a multitude of domains (e.g., academia~\cite{alberola2016artificial}, 
social networking~\cite{anagnostopoulos2012online}, project management~\cite{noll2016global}, healthcare~\cite{nawaz2014teaming}). Building successful teams is a common business strategy (e.g., forming rescue and relief teams in response to an emergency~\cite{gunn2015dynamic}, establishing entrepreneurial teams for new ventures~\cite{lazar2020entrepreneurial}, 
forming teams dynamically in context of multi-agent systems (e.g., supply chains)~\cite{gatson2005agent}). In this paper, we focus on teaming for researchers applying to funding agencies in response to their call for proposals, using group recommendation strategies. The advantage of this setting is that the required data is already publicly available. 
A large amount of research funding in public universities comes from external funding agencies such as National Science Foundation (NSF) and National Institutes of Health (NIH). These opportunities often require multi-disciplinary teams from a wide variety of backgrounds to be quickly assembled. 
However, not every single team is often successful in achieving their goals, due to factors such as lack of accurate information, time, or communication, and incompatibility in terms of skill sets among team members. 

\begin{figure}
    \centering
    \includegraphics[width=1\linewidth]{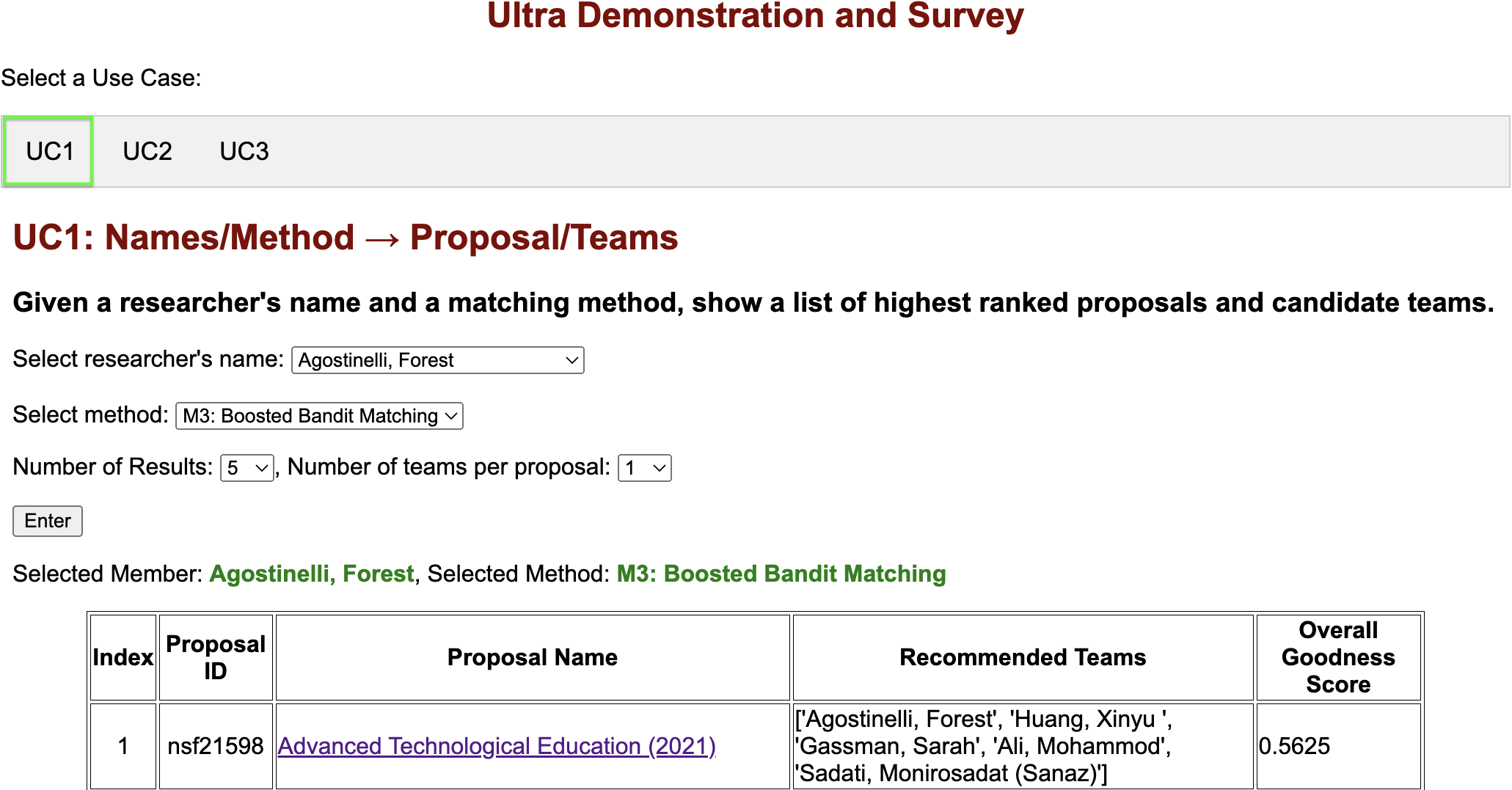}
    \caption{A demo use case, ${UC}_1$, showing a team participant view of ULTRA.}
    \label{fig:ultra_system_usecases}
\end{figure}

We build upon prior work, ULTRA (\textbf{U}niversity-\textbf{L}ead \textbf{T}eam Builder from \textbf{R}FPs and \textbf{A}nalysis)~\cite{srivastava2022ultra}, a novel AI-based prototype for assisting with team formation when researchers respond to calls for proposals from funding agencies. In this paper, we interchangeably use the term \textit{call for proposal} with \textit{request for proposal (RFP).} 
Figure~\ref{fig:ultra_system_usecases} shows a demo\footnote{A full demo interaction with the ULTRA system can be found at https://www.youtube.com/watch?v=8MUtxsfVNIU. The tool is deployed at http://casy.cse.sc.edu/ultra/teaming/. Additional details about usecases, experiments, and survey resources are at \cite{ultra-resources}.} view of how the system works for an individual user who can become a team participant. 
The system first extracts technical skills from proposal calls found at publicly available data sources (e.g., NSF archives) and those present in online profiles of researchers (e.g., personal homepages, Google Scholar history), along with any additional teaming constraints that administrators or team participants of an institution may provide. Using AI and NLP techniques, ULTRA next provides a plausible list of candidate teams for each proposal call, where each team has at least two members. Our prior work~\cite{srivastava2022ultra}, however, is a use case of a {\em sequential} single-item recommendation problem, where solutions are often limited by known issues such as cold start~\cite{abdollahpouri2019managing} or popularity bias~\cite{yalcin2022evaluating}. Therefore, we expand on this work to include group recommendation and novel AI methods to recommend optimal teams. 

Our contributions in the paper are that we:
\begin{itemize}
    \item formulate a novel {\em  group  recommendation problem} to  promote research collaborations where the objective is to suggest teams that could respond to calls for proposals, using skills found in open data, and balancing short- and long-term optimization.
    \item introduce a metric to measure goodness of teams and consider a configurable set of criteria: redundancy, group (set) size, coverage, and robustness.
    \item solve the novel problem using a variety of AI methods: string, taxonomy, and bandit (relational learning) methods, and compare them with a randomized baseline.
    \item  establish the benefit of the solution methods quantitatively using goodness metric. We find that more {\em informed methods lead to recommendations of smaller number of teams but higher goodness}.
    \item  establish the benefit of the system qualitatively using an IRB-approved preliminary survey at a College of a major US University. We find that {\em users broadly consider the tool useful as well as relevant} but more studies are needed. 
    \item demonstrate the generality of the approach with experiments at  two different institutions in US and India.  
    \item create and publish a teaming dataset that is available for research.

\end{itemize}

%% file: sections/relatedwork.tex
\section{Related Work}
A well-studied problem of AI in team formation is the \textit{Hedonic Games} framework~\cite{aziz2017fractional, gairing2010computing}, 
a coalition structure consisting of disjoint coalitions that cover all players, where each player estimates a valuation for other players in their group. Team formation has also been considered in project management, where evaluation of team members is conducted by measuring attributes such as leadership skills, technical talent, problem solving capabilities, and cultural relevance~\cite{
warhuus2021teaming}. Sports leagues assess the physical and functional well-being of players before forming teams~\cite{dadelo2014multi}. Such a task can be complex as 
teams often require players who are efficient in multiple roles. 
\cite{ahmed2013multi} used the NSGA-II algorithm and employed an evolutionary multi-objective approach to obtain a high-performing team for cricket tournaments. Some factors that the system takes into consideration are batting average and bowling performance of individual players, wicket-keeper's past performances, and other rule-based constraints. Depending on the team selection strategy, each of the above factors are ranked (or nulled) according to importance and the final solution set is obtained according to it. If all the factors are deemed equally important, a domination approach is applied instead, i.e., sorting teams according to non-dominating factors (e.g., based on total cost of team formation) and picking the best front solution (i.e., lowest cost team).

Closer to our setting, \cite{machado2019fair} proposed brute force and heuristic approaches to create team recommendations in multidisciplinary projects, where most suitable candidates are incrementally selected until project requirements are fulfilled. However, unlike our work, they do not use any metrics to quantitatively measure the effectiveness of recommended teams, and additionally assume that an individual member may completely fulfill one skill requirement. The complexity of making multi-criteria decisions and building successful teams also increases with the number of available candidates. Evaluating the goodness of every team permutation quickly becomes computationally prohibitive and NP-hard~\cite{roy2014exploiting}. The semantics of group recommendation algorithms were also defined by a means of a Consensus Function, where \textit{disagreement} amongst group members was introduced as a factor to influence the generated recommendations~\cite{ameryahia2009group}. Alternatively, methods have also been proposed to seek a certain level of \textit{agreement} amongst group members, where individuals preferences are iteratively brought closer, until a desired threshold is achieved~\cite{castro2015consensus}. Search-based optimization techniques were also explored using a queuing simulation model to maximize resource usage in software project management~\cite{di2011use}. Another application of team formation is in entrepreneurial domains~\cite{lazar2020entrepreneurial}, where entrepreneurs search for trustworthy partners \textit{and} investors when building new ventures. 

Existing literature systematically answers many questions regarding what, why, and how teams are formed~\cite{costa2020team, juarez2021comprehensive}.
However, recruiting a group of experts to work towards a common goal does not guarantee that they will \textit{always} operate as a team. 

\textbf{Drawbacks of Single-Item Recommendation.} A common objective of many recommender algorithms~\cite{su2009survey, cremonesi2010performance} is to learn each individual's preferences through his interactions with a system, estimate his satisfaction for items he has not interacted with before, and return the top-\textit{k} items with highest estimated ratings. However, many single-item recommender systems suffer from known issues such as popularity bias, 
where suggestions show an uneven distribution in recommending similar items, and the cold start problem, 
a state where a new user wishes to interact with a system but there is not enough information about his preferences to make an accurate decision.

\textbf{Motivation.} We therefore expand on our prior work, ULTRA (\textbf{U}niversity-\textbf{L}ead \textbf{T}eam Builder from \textbf{R}FPs and \textbf{A}nalysis)~\cite{srivastava2022ultra}, and consider a group recommendation setting that promotes research collaboration using novel AI methods to recommend optimal teaming suggestions. Our work encourages multi-functional and interdisciplinary teams to form, and brings together members from various disciplines of work responsibility. 

%% file: sections/problem.tex
\section{Problem Setting}

\subsection{Problem Formulation}

In this section, we introduce the domain of our work and provide the groundwork for the problem. Funding agencies periodically announce Requests For Proposals (RFPs) on specific themes where they are looking for ideas to fund. Researchers in turn respond to those calls, with proposals, where they explain their ideas and detail a meaningful and actionable plan for how they plan to achieve said goals within a specific allotted budget, time frame, and other constraints. In doing so, they also often look to team with other colleagues to respond to such calls. As such, we consider a teaming environment where the availability of candidate participants may change at any given time, often sporadically.

In terms of terminology, let $S$ denote the set of all skills. Then, the {\em demand} for teaming is represented by the set of skills $S_i$ desired by an RFP $c_i$. Further, the  {\em supply}  is represented at an institution by the set of researchers $R$ along with their respective research interests and profiles to satisfy the demand.  
Teaming objectives may be short-term ($ST$), long-term ($LT$), or a combination of the two. An example of a short-term goal would be to meet the immediate requirements and skill needs, $S=\{s_1, s_2, ..., s_\alpha\}$ set forth by a call for proposal $c_i$. Each call in $C=\{c_1, c_2, ..., c_M\}$ requires a very specific skill set $S_i$ and an immediate and optimal assembling of available candidate researchers, $R=\{r_1, r_2, ..., r_N\}$. We then solve our teaming setup in three phases:~(1)~we first \textit{match} all researchers who may be of interest to the calls based on skills needed,~(2)~\textit{group} which subset of researchers should be recommended to be in a team $t_i$, and~(3)~compute goodness score $g_i$ for each team in $T=\{t_1, t_2, ..., t_\beta\}$ and recommend the top-\textit{k} suggestions to interested users. 

Similarly, some example scenarios of a long-term objective would be to maximize the number of funded awards $A$ 
given to a researcher $r_j$ over a  time period (${LT}^A_t$), have a robust (diversified) pool of experienced talent ($LT^R_t$), and satisfy diversity goals of researchers' institutions.

Our system would be of interest to at least two types of users: (1)~administrators at researchers’ organizations (e.g., university institutions) who want to promote more collaborations, proposals, and diversity at their institutions, and (2)~potential team members who will respond to a given RFP $c_i$ and are looking for collaborative opportunities. 
Various environments call for different teams to be formed and matched with relevant opportunities. The candidate member set also may change over time, along with each of their skills and research interests. 


Each user (i.e., admin or researcher) will interact with the system when a call for proposal $c_i$ is announced. Based on the requirements set by $c_i$, along with profiles extracted from researchers, the system will then algorithmically suggest teaming choices $T=\{t_1, t_2, ..., t_\beta\}$, which the users may accept or reject. We evaluate the efficacy of our algorithms and validate the teaming outputs via a goodness score. 

\subsection{Use Cases} \label{subsec:usecases}
We show at least three practical use cases (UCs) to demonstrate the utility of our system. Each use case has various input prompts to select from and includes different algorithms that can be used to recommend or suggest proposals and teams to interested users. 


For the first use case, ${UC}_1$, given a researcher's name and a teaming method $M_i$, we display a list of $k$ highest ranked proposals and possible teams (shown in Figure~\ref{fig:ultra_system_usecases}). Similarly, for ${UC}_2$, given a proposal call $c_i$ from a list of recently announced proposals $C$ (ideally refreshed in real-time or regularly), we display the best possible teams $T$ available for $c_i$. And the final use case, ${UC}_3$, takes input in the form of a research interest and teaming method, and displays respective matching proposals and teams for those parameters. 
We empirically evaluate the three use cases and the functionality of the four methods used in the later sections of the paper. 

%% file: sections/systemdesign.tex
\section{System Design}
In this section, we describe the general architecture of our system and its most important components.

\subsection{Metrics} \label{subsec:metrics}
A challenge in team recommendation scenarios is how to adapt to a group as a whole, given individual preferences of each member within a group~\cite{boratto2010groups}. Literature on team collaboration emphasizes that teams be organized to ensure diversity of team members~\cite{he2022expertise} 
Additionally, when it comes to the relationship between collaboration and scientific impact, team size also matters. 
While equally dependent on the number of requirements and time constraints set forth by an RFP, evidence also suggests that shorter teams also yield quick outputs, 
as they allow for higher accountability, autonomy, and flexibility amongst team members~\cite{trope2023small}. 

In traditional literature on recommendation, there are  metrics to measure the position of a (single correct) result (e.g., mean reciprocal rank (MRR) and top-\textit{k} for ranking). However, our focus is on metrics that reflect the goodness of \textit{multiple} good results (i.e., team recommendations). Still, positional metrics are important since they  reflect users’ acceptance of the results. We have implicitly incorporated them by displaying teaming results in descending order of the team goodness score. Therefore, by incorporating both considerations (team quality and user acceptance), for each candidate team $t_i$, we measure its effectiveness towards $c_i$ using a goodness score $g_i$. The score denotes the chances of success for a team to fulfill the requirements of $c_i$, and is computed by taking the weighted mean of four configurable metrics: \textit{redundancy, set size, coverage,} and \textit{k-robustness.} We now explain the metrics used, along with how goodness is calculated for a team.

\subsubsection{Redundancy.} This is defined as the percentage of demanded skills that are commonly shared amongst multiple researchers. A trade-off that arises with skill redundancy is team robustness versus diversity. While an increased redundancy ensures expert competence within a group of skills, it also risks limiting the amount of skill diversity that a team has.

\subsubsection{Set Size.} This metric is defined as the size of the candidate team. A trade-off present here is skill coverage and robustness versus funding amount split per researcher. A smaller team size runs the risk of not being able to accomplish the necessary goals of the proposal call (e.g., due to unavailability of any team members, longer efforts needed to complete a task), whereas a larger one mitigates that risk but lowers the amount of funding that each researcher receives. In addition, large teams lead to other disadvantages such as unequal participation amongst members, longer time needed to make decisions (e.g., due to intrinsic conflicts or lack of timely communication), whereas a smaller team fosters for accountability, individuality, and flexibility in ideas and schedules~\cite{mcleod1996ethnic}.

\subsubsection{Coverage.} This metric represents the percentage of proposal-required skills that are satisfied by the candidate team as a whole. A candidate member's skills are defined from those listed within their personal webpages and extracted from other research profiles such as Google Scholar. We devise this metric by borrowing the idea that diverse expertise often invites individuals with different perspectives but also may lack common shared experience~\cite{he2022expertise}. A mix of team members with diverse knowledge, skill sets, and abilities have also been thought to bring forth their unique skills to the team and provide it with the broadest possible skill set. 

\subsubsection{\textit{k}-Robustness.} We borrow this metric from~\cite{okimoto2015form}, where each team needs to be able to equally satisfy the teaming constraints even after the removal or unavailability of \textit{k} researchers. Such a team is defined as \textit{k}-robust. 

\subsubsection{Goodness Score.} We first normalize the aforementioned metrics to make their values query-independent. Next, we assign each of the metrics a predefined weight by default. Given our use cases, the weights are defined by the intuition to yield the maximum profit (i.e., project completion) and credibility (i.e., project quality) a team can achieve. A high coverage and robustness are therefore more desired for overall project success, whereas high redundancy and set size are less prioritized. As a result, for each candidate team $t_i$, we penalize the latter two metrics and reward the former. The penalized metrics are set to a negative weight of $-1$, whereas the desired ones are set to a positive weight of $+1$. Finally, the goodness score for a team is calculated using the weighted mean from all four metrics. The diversity of metrics increases the potential to provide strength and resilience to the overall model. For additional reference, we make our metrics tool publicly available on GitHub~\footnote{Metrics tool: https://github.com/ai4society/Ultra-Metric}.

\subsection{Methods}
\subsubsection{M0: Random Team Formation.}
\begin{algorithm} 
	\caption{{\it M0:} Random Team Formation} 
 {\small
 \begin{flushleft}
        \hspace*{\algorithmicindent} \textbf{Input:} Calls $C=\{c_1, \ldots, c_M\}$, Researchers $R=\{r_1, \ldots, r_N\}$\\
        \hspace*{\algorithmicindent} \textbf{Output:} Teams $T=\{t_1, \ldots, t_\beta\}$, good. scores
        $G=\{g_1, \ldots, g_\beta\}$
 \end{flushleft}
 \label{algo:m0}
	\begin{algorithmic}[1]
		\For {$c_i \in C$}
            \State $T$ = \{\}, $G$ = \{\}     
                \For {$j=1,2,\ldots,\beta$}
            	\State Let $t_j$ be a \textit{k} random sampling of $N$ researchers.
            \State $T$ = $T \bigcup \{t_j\}$ 
            \State Compute goodness score $g_j$ for team $t_j$. 
            \State $G$ = $G \bigcup \{g_j\}$ 
		      \EndFor
		\EndFor
	\end{algorithmic} 
 }
\end{algorithm}

We consider random team selection as our baseline, where candidate teams are formed in a randomized manner, without any adherence to the skills in demand, and matched to an arbitrary proposal, regardless of relevance. Given any $c_i$, we select a random number of individuals from a pool of $N$ available researchers to form teams $T$. Using our goodness function, we next evaluate each team $t_j$ by extracting the skills $S_i$ required by the proposal $c_i$ and checking how many are solvable by the team at hand. Algorithm~\ref{algo:m0} provides pseudocode for {\it M0}.

\subsubsection{M1: Team Formation Using String Matching.}
Given a call for proposal $c_i$, we extract the technical skills required from it using its title and synopsis as inputs. We remove stop words and delimiters and use keyword extraction to gather only the relevant skills needed in $S_i$. Similarly, we gather the research interests each available researcher $r_j$ has listed on their personal webpage and demonstrated with their Google Scholar history. We denote this as $\sigma(r_j)$. We then check if there are any common interests between $\sigma(r_j)$ and $S_i$. 

\begin{algorithm} 
	\caption{{\it M1:} Team Formation Using String Matching} 
 {\small
 \begin{flushleft}
        \hspace*{\algorithmicindent} 
        \textbf{Input:} Calls $C=\{c_1, \ldots, c_M\}$, Researchers $R=\{r_1, \ldots, r_N\}$, string matching threshold $th_{M1}$\\
        \hspace*{\algorithmicindent} 
        \textbf{Output:} Teams $T=\{t_1, \ldots, t_\beta\}$, good. scores
        $G=\{g_1, \ldots, g_\beta\}$
 \end{flushleft}
 \label{algo:m1}
	\begin{algorithmic}[1]
		\For {$c_i \in C$}
              \State $T$ = \{\}, $G$ = \{\}  
                \State Extract technical skills $S_i=\{s_1, s_2, \ldots, s_\alpha\}$ for each $c_i$.
                \State Initialize $candidate\_researchers=[]$.
                \For {$s \in S$}
                    \For {$j=1,2,\ldots,N$}
                        \If {$s$ is in $\sigma(r_j)$ by satisfying $th_{M1}$}
                      \State Add $r_j$ to $candidate\_researchers~[]$. 
                      \EndIf
                    \EndFor
                \EndFor
            \State Using $candidate\_researchers~[]$, form each team $t_k$, prioritizing members with highest string matches.
            \State $T$ = $T \bigcup  \{t_k\}$ 
            \State Compute goodness score $g_k$ for each team $t_k$. 
            \State $G$ = $G \bigcup  \{g_k\}$ 

		      \EndFor
	\end{algorithmic} 
 }
\end{algorithm}
\begin{algorithm} 
	\caption{{\it M2:} Team Formation Using Taxonomical Matching} 
 {\small
 
 \begin{flushleft}
        \hspace*{\algorithmicindent} 
        \textbf{Input:} Calls $C=\{c_1, \ldots, c_M\}$, Researchers $R=\{r_1, \ldots, r_N\}$, string matching threshold $th_{M2}$\\
        \hspace*{\algorithmicindent} 
        \textbf{Output:} Teams $T=\{t_1, \ldots, t_\beta\}$, good. scores
        $G=\{g_1, \ldots, g_\beta\}$ 
 \end{flushleft}
 \label{algo:m2}
	\begin{algorithmic}[1]
		\For {$c_i \in C$}
                \State $T$ = \{\}, $G$ = \{\}     
                \State Extract technical skills $S_i=\{s_1, s_2, \ldots, s_\alpha\}$ for each $c_i$.
                \State Calculate \textit{n-grams} from $c_i$, and add to $S_i$.
                \State {Using $th_{M2}$, map each $s \in S_i$ with relevant classification codes from ACM-CCS, i.e., $\delta(S_i)$.}
                \State Initialize $candidate\_researchers=[]$.
                \For {$j=1,2,\ldots,N$}
                    \State {Using the same $th_{M2}$, calculate $\delta(\sigma(r_j)$.}
                    \If {$\delta(S_i) \bigcap \delta(\sigma(r_j) != \emptyset$}
                        \State Add $r_j$ to $candidate\_researchers~[]$.
                    \EndIf
                \EndFor

            \State Using $candidate\_researchers~[]$ and for each $r_j$, form each team $t_k$, prioritizing members with highest taxonomical matches.
            \State $T$ = $T \bigcup  \{t_k\}$ 
            \State Compute goodness scores $g_k$ for each team $t_k$. 
            \State $G$ = $G \bigcup  \{g_k\}$ 
        \EndFor
    \end{algorithmic} 
    }
\end{algorithm}
Given a pattern string of length \textit{x} and a target string of length \textit{y}, we determine if there is any overlap between those two using a string-match threshold $th_{M1}$. Algorithm~\ref{algo:m1} provides pseudocode for {\it M1}.

\subsubsection{M2: Team Formation Using Taxonomical Matching.}
We further improve the accuracy and precision of the previous methods, {\it M0} and {\it M1}, by considering query-based \textit{semantic matching}, combined with the use of a \textit{taxonomy}. We use a poly-hierarchical, subject-based ontology, provided by the \textit{ACM Computing Classification System (CCS)}~\cite{acm2012acm}. There are over two thousand topics listed that broadly reflect the research areas pursued in the computing discipline. These are further organized into categories and concepts, with up to four branches of structure. We use this ontology to determine if two research skills may be matched semantically rather than only string-wise. For instance, if two researchers, $r_a$ and $r_b$, each had the respective skills, \textit{``natural language processing"} and \textit{``knowledge representation"}, the method {\it M1} will return an extremely low string-match score and deny any association between the terms. However, using ACM-CCS, {\it M2} will categorize these two interests under \textit{``artificial intelligence"} and consider the possibility that $r_a$ and $r_b$ may belong within the same team for a call $c_i$. 

Given a call for proposal $c_i$, we extract the relevant skills needed, $S_i$. For each skill in $S_i$, we then use \textit{n-grams} to compare each sequence of words with the concepts in ACM-CCS using a string match threshold $th_{M2}$. We denote this step with $\delta(S_i)$. Each concept is mapped to a specific code, and we use those codes to search for candidate researchers. For each available researcher $r_j$, we first extract their research interests $\sigma(r_j)$. For each interest, we similarly calculate $\delta(\sigma(r_j))$ to get the relevant codes from ACM-CCS. Finally, we form teams and calculate goodness based on a match between the codes. Algorithm~\ref{algo:m2} provides pseudocode for {\it M2}.

\subsubsection{M3: Team Formation Using Boosted Bandit.}
According to requirements of proposals and expertise of researchers, the previous three methods apply manually-crafted rules to matching researchers to teams. However, {\it M3} extracts rules automatically from data. With more facts and data provided, this method is able to learn more complex rules automatically. We consider the strategy taken by \cite{kakadiya2021relational}, and formulate the problem as team recommendation using contextual bandits. The key idea is that given the skills required by proposals and the research interests/expertise (denoted as $\mathbf{x}$), 
\begin{algorithm} 
\caption{{\it M3:} Team Formation Using Boosted Bandit} 
{\small
 \begin{flushleft}
        \hspace*{\algorithmicindent} 
        \textbf{Input:} Calls $C=\{c_1, \ldots, c_M\}$, Researchers $R=\{r_1, \ldots, r_N\}$\\
        \hspace*{\algorithmicindent} 
        \textbf{Output:} Teams $T=\{t_1, \ldots, t_\beta\}$, good. scores
        $G=\{g_1, \ldots, g_\beta\}$
 \end{flushleft}
 \label{algo:m3}
	\begin{algorithmic}[1]
        \Function{BoostedTrees}{$predicates$}
            \For {$p \in predicates$}
            \State Let $I$ be a pre-set number of iterations.
                \For {$i=1,2,\ldots,I$}
                \State Generate examples for the
regression-tree learner: \Call{GenerateEx}{$p, predicates, F^p_{i-1}$}
                \State Get new regression tree, which approximates functional gradient $\triangle_i(p)$, and update current model $F^p_i$.
                \EndFor
            \State Get final potential: $\psi_I=\psi_0+\triangle_1(p)+\ldots+\triangle_I(p)$
            \EndFor
        \State \Return
        
        \EndFunction
            \Function{GenerateEx}{$p, predicates, F$}
                \State Initialize examples $E=\emptyset$.
                \For {$j=1,2,\ldots,X_p$} 
                    \State Calculate probability of predicate $p$ being true.
                    \State Compute gradient and update regression examples.
                    \State Compute regression values based on the groundings of current example.
                \EndFor
            \State \Return regression examples $E$.
            \EndFunction
		\For {$c_i \in C$}
                \State $T$ = \{\}, $G$ = \{\}     
                \State Extract technical skills $S_i=\{s_1, s_2, \ldots, s_\alpha\}$ for each $c_i$.
                \State Initialize $candidate\_researchers=[]$.
                \State Initialize $predicates[]$.
                \State Show associations between ($c_i$, $S_i$), and ($r_j$, $\sigma(r_j)$) as $predicates[]$.
            \State Get candidate researchers from \Call{BoostedTrees}{$predicates$} and add to $candidate\_researchers~[]$. Form each team $t_k$, prioritizing members with highest probability matches.
            \State $T$ = $T \bigcup  \{t_k\}$ 
            \State Compute goodness score $g_k$ for each team $t_k$. 
            \State $G$ = $G \bigcup  \{g_k\}$ 
        \EndFor
    \end{algorithmic} 
    }
\end{algorithm}  the goal is to learn $P(y\mid\mathbf{x})$ where $y$ is whether a researcher is a potential candidate for the proposal. This is to say that $y$ is a two-argument predicate $candidate(r_j,c_i)$, which states that the researcher $r_j$ is a candidate for the proposal $c_i$. The key idea in boosted bandit is to represent this as a sigmoid, $P(y\mid\mathbf{x}) = \frac{e^{ \psi(y\mid\mathbf{x})}}{\sum_{y' }e^{ \psi(y'\mid\mathbf{x})}}$ and boost this using the machinery of gradient-boosting~\cite{friedman2000additive,dietterich2004training}. Since our data is naturally relational, we adapt the relational boosted bandits for this case~\cite{kakadiya2021relational}.

We have $m$ regression trees for each predicate $p$, where $m$ is the number of time steps or iterations. Each iteration approximates the corresponding gradient for $p$, and each of the trees serve as individual components for the final potential function~$\psi$. The algorithm \Call{BoostedTrees}{$predicates$} then loops across all predicates and learns the potentials for each. The set of regression trees for each predicate then forms the structure of the conditional probability distribution and the set of leaves of each tree form the parameters of the conditional distribution. 

Algorithm~\ref{algo:m3} provides the pseudocode for {\it M3}. We first represent all data in the form of predicates. There are three types of relationships: (1)~every call for proposal $c_i$ mapped to a skill set $S_i$, (2)~every researcher $r_j$ mapped to their research interests $\sigma(r_j)$, and (3)~every call for proposal $c_i$ teamed with a group of researchers $R$ according to the demand and supply. For each predicate $p$ (shown in \Call{BoostedTrees}{$predicates$}), we generate multiple examples $E$ for our regression-tree learner to get new regression trees and update the current model $F^p_i$ at every iteration $i$. The function \Call{GenerateEx}{$p, predicates, F^p_{i-1}$} iterates over all the examples $E$ and computes probability and gradient for each. These probabilities are later used to form teams using a greedy policy, where candidate members with highest probabilities are prioritized first when forming teams.

\subsection{ULTRA System and Survey Deployment}
We built a UI for our system and deployed it using the three use cases detailed in the previous section. 
ULTRA consists of a three-layered architecture: (1)~data storage and retrieval, (2)~team matching, and (3)~analysis of results. (1)~The data we used to perform our experiments is publicly available: (a)~calls for proposals from NSF archives, and (b)~faculty directories at our university. These are retrieved and stored in a separate database, where they are periodically refreshed to get the latest information. (2)~We use the input data and a matching method to view teaming results, along with their respective goodness scores. (3)~We evaluate the results both computationally and empirically. 

We measure the quality of teaming suggestions, and user satisfaction by conducting a user study for ULTRA
over a span of 28~days. We invited researchers from a college within our university to explore the tool and assess its functionality and satisfaction using a feedback survey for every result. The survey includes two 5-point Likert Scale questions: (1) \textit{How relevant is the output given the input?}, and (2) \textit{How useful is the output?}. We also provide a freeform section for additional comments, if any. 

%% file: sections/evaluation.tex
\section{Evaluation} \label{sec:evaluation}

\begin{table}
    \centering
    \begin{tabular}{c|c|c}
        \toprule 
    Method  & Average Quality & Average Volume \\
    \midrule 
    \textit{M0} & $0.0879\pm0.0290$ & $10$ \\
    \textit{M1} & $0.3673\pm0.1569$ & $10$ \\
    \textit{M2} & $0.4097\pm0.1313$ & $9$ \\
    \textit{M3} & $0.5295\pm0.0816$ & $6$ \\
    \bottomrule
    \end{tabular}
    \caption{Average quality ($G$) and volume of teams (\#$T$) shown \textit{per} researcher ($r_j$) at USC. This was done for each method $M_i$, across 434 RFPs and 200 researchers. For average quality, we report the mean and standard deviation, denoted as \texttt{mean$\pm$STD}.} \label{tab:volume_quality}
\end{table}

\subsection{Computational Evaluation of Output}
\subsubsection{Quality vs. Volume of Teams.}

For each method, we assess the quality (goodness) and volume (size) of each teaming suggestion that every researcher $r_j$ receives per every call for proposal $c_i$. Our experiments iterate across a dataset of 434 RFPs and 200 researchers. For each call, every researcher has a maximum cutoff of 10 teams. For each method, we then find the average goodness ($G$) of teams $(T)$ a researcher $r_j$ has been recommended.
The more advanced the method, the better teams a researcher receives. We observe another unique trade-off as a result, where teams formed algorithmically led to an increased precision and quality, and a notable decrease in quantity. \textit{M0}, random team formation, showed poor quality in results with an average goodness of only $0.0879,$ despite the number of teams being abundantly available. \textit{M3}, on the other hand, shows a decrease in the number of teaming choices available for a researcher, but a visible increase in quality, compared to the average goodness for \textit{M0}.
Table~\ref{tab:volume_quality} shows the overall results. 

\subsection{Human Evaluation of Output} 

We deployed our tool at a college-wide level and requested participation from faculty members. This research study has been IRB-approved by our university (IRB\#~Pro00127449) and reflects an observational (unmonitored) qualitative analysis, where we make the tool publicly available and give participants a demo of its usage but do not actively control their actions. This helps us receive many responses quickly and at a low cost, albeit not without its own limitations. For instance, it does not require users to explore all use cases and methods. Therefore, we only make inferences from data that is recorded and do not draw any from those that are left out. As a possible future work, we can perform a controlled qualitative analysis, where we request participants to come to a designated lab, try every pathway, and give feedback. This will then enable us to answer how a user compared the effectiveness of each method on the same example. 

\begin{table}[ht]
{\renewcommand\arraystretch{1.25}
{\begin{tabular}{l|l|l} \hline
& \multicolumn{2}{l}{Qualitative Results} \\ \hline
Comments & \multicolumn{2}{p{6cm}}{1. \raggedright \textit{``This is incredible and has a lot of potential. Can't wait for this to be in real time!"}} \\ 
 & \multicolumn{2}{p{6cm}}{2. \raggedright \textit{``Very well thought out! Great resource to the university"}} \\ 
 & \multicolumn{2}{p{6cm}}{3. \raggedright \textit{``Lots of new people to choose from here! Great work!"}} \\ 
 & \multicolumn{2}{p{6cm}}{4. \raggedright \textit{``Very useful tool  overall, I could see the practical usage of this work!"}} \\ 
 & \multicolumn{2}{p{6cm}}{5. \raggedright \textit{``Would love an explanation for all of your methods used!!"}} \\ 
 \hline
Feedback & \multicolumn{2}{p{6cm}}{1. \raggedright \textit{``Seeing many team pitches here from interdisciplinary domains. Can we choose from say, two, settings where we may choose to work with those from a similar domain or a different one? And how so is the overall goodness score calculated?"}} \\ 
 & \multicolumn{2}{p{6cm}}{2. \raggedright \textit{``A search bar would be great for this one, not just a dropdown!"}} \\ 
 & \multicolumn{2}{p{6cm}}{3. \raggedright \textit{``In addition to the proposal, can we also add research interest as a user-given input? As a merging of the second and third use cases"}} \\ 
 & \multicolumn{2}{p{6cm}}{4. \raggedright \textit{``can we build our own teams for a grant?"}} \\ 
 
\hline
\end{tabular}}}
\caption{A sample of the comments as well as feedback for improvement received from the human study.} \label{tab:comments_feedback}
\end{table}

\begin{figure*}
    \centering
    \includegraphics[width=0.85\textwidth]{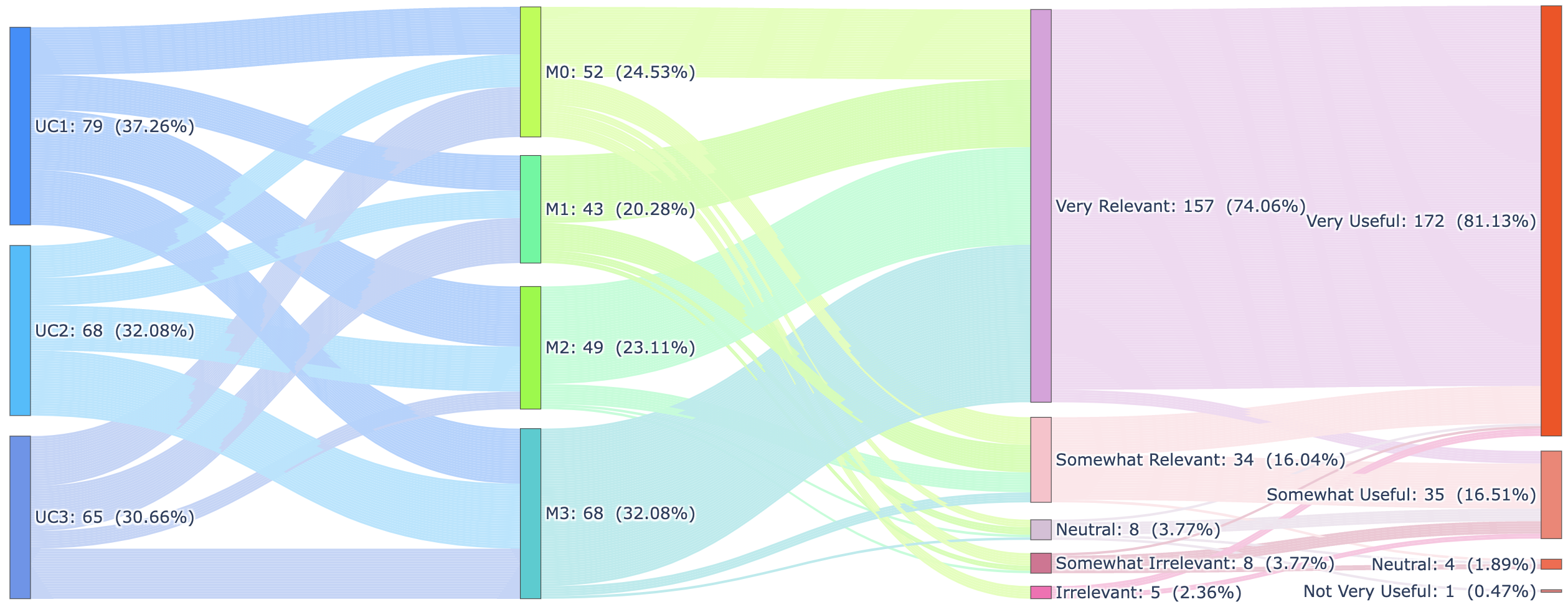}
    \caption{Sankey breakdown of the 212 responses we received from our human study. It shows four categories of nodes: (1)~use cases, (2)~methods, (3)~scale of relevancy, and (4)~scale of usefulness. Each line maps a connection from node type to another, showing the flow of interaction that users had with ULTRA. }
    \label{fig:sankey_human_study}
\end{figure*}

We received a total of 212 responses. Regarding relevancy of outputs, 157 answers were rated as \textit{very relevant} and 34 as \textit{somewhat relevant}, summing to 90.09\% of all responses. Similarly, in terms of tool utility satisfaction, 172 answers responded with \textit{very useful} and 35 with \textit{somewhat useful}, totaling 97.64\% of all responses. Figure~\ref{fig:sankey_human_study} shows a more quantitative breakdown of the responses and Table~\ref{tab:comments_feedback} highlights a few comments and feedback we received. 

\label{subsec:humanstudy}

We observe two repeated patterns of responses regarding \textit{M0}: (1) Teaming choices that were rated as \textit{`somewhat irrelevant'} or \textit{`irrelevant'}, yet still \textit{`very useful'}, and (2) teaming choices that were rated as \textit{`very relevant'}, despite the irrelevancy in the results. Upon analyzing the comments, some mentioned that \textit{M0} could regardless be useful in working with new colleagues from other departments and \textit{`expand new connections'} as a result. Furthermore, we have only requested users to judge the \textit{usability} and \textit{functionality} of our tool, not immediate applicability. They may also be unaware of each candidate member's skill set, which impedes on their ability to accurately quantify a team's success. Due to that, there is a need and role of AI in teaming, as reflected in comments as well. We also factor this intuition into the interpretation of results.

%% file: sections/discussion.tex
\section{Discussion - \textit{Ultra as a Deployed Application and Its Generality}}
In this section, we discuss the characteristics of ULTRA as required for a deployment track paper but not discussed elsewhere. We describe its development experience leading to deployment at a US university. We also demonstrate its generality by considering an altogether different setting from India but leading to similar results: with increased sophistication of methods, the quality of teams increase and the amount of recommended teams decrease. 

\subsection{Development and Deployment}
The development of ULTRA first began in Summer 2021 with a team of 8 developers and an initial prototype was created within 3 months with M0 and M1 for user feedback (August 2021). It included a very small dataset of researchers and RFPs, a single method (i.e., a string-based matching algorithm and greedy teaming strategy), no goodness score and instead only a match percentage threshold, and a smaller-scale empirical evaluation with a few experts at a single University that indicated the promise of such a tool. Once the pilot study results seemed promising, ULTRA was refreshed with additional data; re-imagined and re-designed in Fall 2022; improved through iterative review, feedback, and testing; enhanced with \textit{M2} and \textit{M3,} and alpha-tested with users from different departments at the University of South Carolina for a year. The system was deployed on May 1, 2023 and evaluated under an IRB-approved protocol for 28 days, and the results are as reported in the paper. It remains publicly available as of writing this paper (November 2023). 
During July-August 2023, we evaluated ULTRA with data from another setting in a different country: researchers at Indian Institute of Technology-Roorkee (IIT-R), India and calls from India's funding agencies.   

One main challenge during development was access to clean data related to RFPs and relevant researcher profiles. Due to inconsistencies in data formatting, missing values, and irrelevant or out-of-date entries, exploring automated approaches had been unsuccessful, and data cleaning had  only been possible through iterative collaborations. 
After deployment, additional challenges were raised. One challenge was maintaining data integrity across changes to ULTRA's servers and infrastructure, and another was change in researchers over time due to hiring or attrition. It was essential to run frequent tests to enhance user experience, create a working feedback loop, and ensure a scalable architecture with minimal overhead. 

\begin{table}
    \centering
    \begin{tabular}{c|c|c}
    \toprule 
    Method  & Average Quality & Average Volume \\
    \midrule 
    \textit{M0} & $0.0896\pm0.0006$ & $10$ \\
    \textit{M1} & $0.4218\pm0.0011$ & $8$ \\
    \textit{M2} & $0.4292\pm0.0017$ & $7$ \\
    \textit{M3} & $0.5835\pm0.0203$ & $1$ \\
    \bottomrule
    \end{tabular}
    \caption{Average quality ($G$) and volume of teams (\#$T$) shown \textit{per} researcher ($r_j$) at IIT-R. This was done for each method $M_i$, across 100 RFPs and 46 researchers. For average quality, we report the mean and standard deviation, denoted as \texttt{mean$\pm$STD}.} \label{tab:volume_quality_iitr}
\end{table}

\subsection{Generalizing to Second Institution: IIT-R}
We additionally extended ULTRA to another university in a different region of the world: Indian Institute of Technology-Roorkee (IIT-R), India. We gathered publicly available data on RFPs from the Department of Science \& Technology (DST), a division of Research and Development (R\&D) programmes belonging to and funded by the Government of India's Ministry of Science and Technology~\cite{dst2023department}. We further collected data about IIT-R's faculty members and their respective research profiles. 
Our initial evaluation is with 100 RFPs and 46 researchers.
Table~\ref{tab:volume_quality_iitr} shows the computational evaluation of ULTRA on IIT-R's data. 
From the quantitative results, we observe a similar trend to Table~\ref{tab:volume_quality}, where teams formed algorithmically led to an increased precision and quality, and visible decrease in quantity.
Since human study at any institution is subject to local policies and workplace culture, we leave performance of such a study at IIT-R as a future work.

%% file: sections/conclusion.tex
\section{Conclusion}
To conclude, we presented the problem of building teams for funding that allows for collaboration opportunities. We then created and implemented  AI methods using string, taxonomy, and more advanced contextual boosted bandits in ULTRA, and demonstrated them to be quite useful both quantitatively (where informed methods increase recommendation quality while reducing their volume) and qualitatively in real human evaluations. 
We showed the generality of our approach in two different settings from US and India, and discussed our experience of deploying the system.

One area to extend our work is by considering larger data sizes for both researchers and RFPs, and from more diverse  sources. Furthermore, since our methods are dependent on open data about researchers as well as proposal calls, this dependency can be both a source of strength and weakness. Data has the potential to encourage  teaming without human bias but can also lead to inferior recommendation if the data is obsolete.  
Similarly, any feedback on the success of recommendation is only possible when a proposal has been won, and this data is usually not available or quite delayed (months or years after recommendation) to be useful. Considering a longer time frame for recommendation with proposal success data could lead to better results. Another area is to further enhance our goodness score with more metrics (e.g., considering the relevance of a researcher's previous projects to current RFPs and the number of ongoing projects a researcher is engaged in).


Our work inspires several interesting future extensions: considering a variety of domain knowledge including but not limited to fairness constraints, teaming constraints, domain constraints, etc. and developing a knowledge-driven learning system that can both exploit both the data and such knowledge remains an exciting future direction. A second direction would be to develop methods that would allow for interactive teaming where the system could not only present the recommendations but explain why and allow for human inputs to be used for refining the learned models. In addition, the scale of our survey could also be improved to ask about more aspects of the teams from participants: diversity and the connection strength between members, etc. A final and important direction is to scale this to different collaborative settings including but not limited to: healthcare, finance, law/legal, mental health, and educational support. 

%% file: sections/acknowledgement.tex
\section*{Acknowledgements}
We would like to thank Michael Widener,  Sugata Gangopadhyay and Sandeep Kumar for their valuable assistance in the data collection and processing for IIT-R; Rohit Sharma and Owen Bond for an earlier version of ULTRA; and Michael Huhns, Danielle McElwain, Michael Matthews, and  Paul Ziehl for discussions. Siva Likitha Valluru, Biplav Srivastava, and Sai Teja Paladi acknowledge funding from South Carolina Research Agency, Cisco Research, and VAJRA program, while Siwen Yan and Sriraam Natarajan acknowledge AFOSR under award FA9550-19-1-0391.